\pdfoutput=1
\documentclass[11pt]{article}

\usepackage[preprint]{acl}

\usepackage{times}
\usepackage{latexsym}

\usepackage[T1]{fontenc}
\usepackage[utf8]{inputenc}

\usepackage{microtype}

\usepackage{inconsolata}

\usepackage{graphicx}

\usepackage{amsmath}
\usepackage{amssymb}
\usepackage{mathtools}
\usepackage{amsthm}
\usepackage{amsfonts}
\usepackage{multirow}
\usepackage{booktabs}
\usepackage{makecell}
\usepackage[most]{tcolorbox}
\usepackage{algorithm}
\usepackage{algorithmic}

\newcommand{\ModelName}{{CoCT}}

\title{Chain-of-Conceptual-Thought Elicits Daily Conversation in Large Language Models}

\author{
    \textbf{Qingqing Gu\textsuperscript{1}},
    \textbf{Dan Wang\textsuperscript{1}},
  \textbf{Yue Zhao\textsuperscript{1}},
  \textbf{Xiaoyu Wang\textsuperscript{1,2}\thanks{Work was done during the internship at Geely.}}, \\
  \textbf{Zhonglin Jiang \textsuperscript{1}},
  \textbf{Yong Chen\textsuperscript{1}},
  \textbf{Hongyan Li\textsuperscript{1}},
  \textbf{Luo Ji\textsuperscript{1}\thanks{Corresponding Author.}}
\\
\\
  \textsuperscript{1} Geely AI Lab, Beijing, China \textsuperscript{2} Beijing Institute of Technology, Beijing, China
\\
  \small{
    \textbf{Correspondence:} \href{mailto:email@domain}{Luo.Ji1@geely.com}
  }
}

\begin{document}

\maketitle

\begin{abstract}
Chain-of-Thought (CoT) is widely applied to enhance the LLM capability in math, coding and reasoning tasks. However, its performance is limited for open-domain tasks, when there are no clearly defined reasoning steps or logical transitions. To mitigate such challenges, we propose a new prompt-based paradigm called Chain of Conceptual Thoughts (CoCT), which suggests the LLM first to produce the tag of concepts, then complete the detailed content following the concept. To encourage this hierarchical way of thinking, we implement the concepts with emotions, strategies and topics. We experiment with this paradigm in daily and emotional support conversations, covering tasks with both in-domain and out-of-domain concept settings. Automatic, human, and LLM-based evaluations reveal that CoCT surpasses several prompt-based baselines such as self-refine, ECoT, SoT and RAG, suggesting a potential solution of LLM prompting paradigm for a wider scope of tasks.
\end{abstract}

\section{Introduction}
\label{sec:introduction}

Large Language Models (LLMs) have demonstrated remarkable capabilities in tasks of natural language understanding, question-answering, daily conversation, and complicated reasoning \cite{Ouyang2022ChatGPT,llama3modelcard}.To further elicit their reasoning abilities, a Chain-of-Thought (CoT) paradigm \cite{wei2022chain} is typically adopted, which prompts the LLM to think `step by step' before producing the final response. Several effective paradigms are then proposed by incorporating CoT with specific reasoning hierarchies such as Tree of Thoughts \cite{yao2023tree}, Graph of Thoughts \cite{10.1609/aaai.v38i16.29720}, Structured Chain-of-Thought \cite{10.1145/3690635}, or with test-time scaling \cite{ac1f09077393404a8bea5141d8710259,10.5555/3666122.3666499,Madaan2023SelfRefine,ning2024skeletonofthought}.

%, and complicated reasoning
%iCoT \cite{deng2023implicitchainthoughtreasoning}

\begin{figure}[!t]
\centering
  \includegraphics[width=1.0\linewidth]{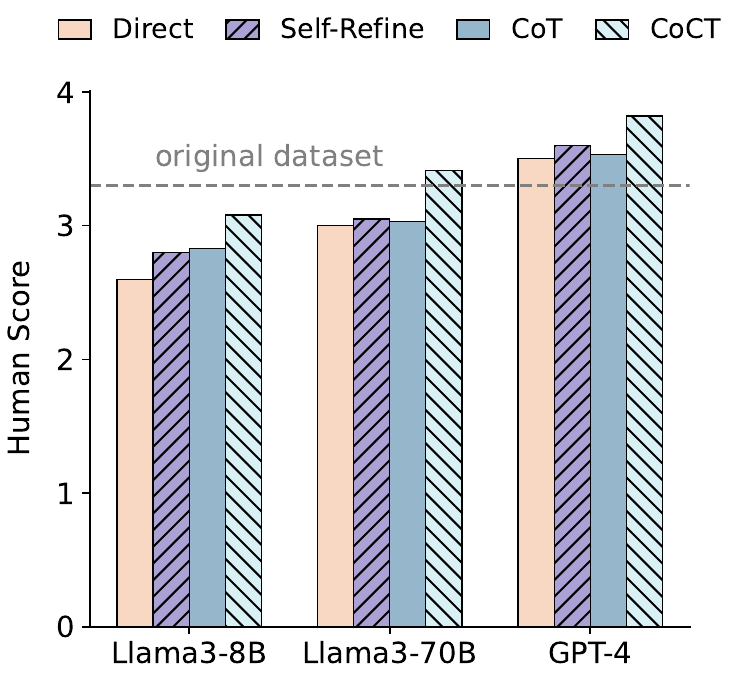} 
  \caption{Comparison between {\ModelName} and other prompt-based baselines on ESconv.}
  \label{fig:result_snapshot} 
\end{figure}

Although effective, such CoT-related methods are often focused on math problems, code generation, or other reasoning scenarios that have clear reasoning steps. When dealing with open-domain tasks such as conversation, CoT (or CoT-like methods on decision trees or graphs) \cite{wei2022chain,yao2023tree,10.1609/aaai.v38i16.29720,10.1145/3690635} may encounter difficulties since conversations generally lack clear definitions of reasoning steps, step transitions, and stepwise standardized answers. To alleviate such issues, ECoT \cite{li2024enhancingemotionalgenerationcapability,kang-etal-2024-large} is proposed which encourages sequential emotional decisions on emotional support conversations (ESC). SoT \cite{ning2024skeletonofthought} instead first determines the skeleton tokens, then generates the completed response. However, methods such as ECoT and SoT still do not provide solutions to open-domain conversations and also introduce excessive computational costs from iterative inference.

%Another type of method leverages the multi-hop inference to self-refine \cite{Madaan2023SelfRefine} the original response, complete the skeleton response \cite{ning2024skeletonofthought} or response enhancement by self-play \cite{zhou-etal-2024-think}. 

%SoT \cite{ning2024skeletonofthought} instead first determines the skeleton tokens then generates the completed response. However, methods such as ECoT and SoT still do not provide solutions to open-domain conversations and also introduce excessive computational costs from iterative inference.

%LLMs with CoT prompts might encounter performance degradation. For such scenarios, there is no clear definition about the reason steps, the transition between different steps, as well as the standard answer of each step. CoT variants based on more structured modules (\textit{e.g.}, decision tree, graph) also become intractable for such scenarios. Such issues hinder the application of LLMs on a broader scope.

\begin{table*}[htbp!]
\caption{Lists of annotated emotions, strategies and topics from ESConv \cite{liu2021ESconv}, DailyDialogue \cite{li-etal-2017-dailydialog}, EmpatheticDialogues \cite{rashkin-etal-2019-towards} and Cskills \cite{zhou-etal-2024-think}, respectively. Strategies of ESconv are abbreviated (for full names, refer to \cite{liu2021ESconv}). Only the most frequent 10 emotions of EmpatheticDialogues and the most frequent 8 topics of Cskills are listed.}
\label{tab:dataset_concepts}
\centering
%\small
\resizebox{\textwidth}{!}{
%\setlength{\tabcolsep}{8pt} % 调整列间距
%\begin{tabular}{>{\centering\arraybackslash}m{0.25\textwidth}|p{0.7\textwidth}}
\begin{tabular}{c | c | l}
%\Xhline{1.2pt}
\toprule
\multicolumn{1}{c|}{Dataset} & \multicolumn{1}{c|}{Concepts} & \multicolumn{1}{c}{Concept Categories}  \\ 
\toprule
\multirow{2}[0]{*}{ESconv} & \textit{Emotion} & anger, anxiety, depression, disgust, fear, nervousness, sadness, shame \\ 
\cmidrule{2-3}
 & \textit{Strategy} & Que., Paraphrasing \&Res., Ref., Self-Dis., Aff.\& Rea., Pro., Inf., Others \\ 
\midrule
\multirow{2}[0]{*}{DailyDialgoue} & \textit{Emotion} & anger, disgust, fear, happiness, sadness, surprise, no emotion \\ 
\cmidrule{2-3}
 & \textit{Strategy} &inform, question, directive, and commissive \\ 
 \midrule
EmpatheticDialogues & \textit{Emotion} & surprised, grateful, proud, sentimental, annoyed, excited, sad, disgusted, $\dots$ \\ %, angry, joyful
\midrule
Cskills & \textit{Topic} &  sports, travel, art, music, technology, food\&drink, hobbies, entertainment, $\dots$ \\ %\&crafts, \&animal % -ConceptGuidence
%\Xhline{1.2pt}
\bottomrule
\end{tabular}
}
\end{table*}

\begin{figure*}[!t]
\centering
  \includegraphics[width=1\linewidth]{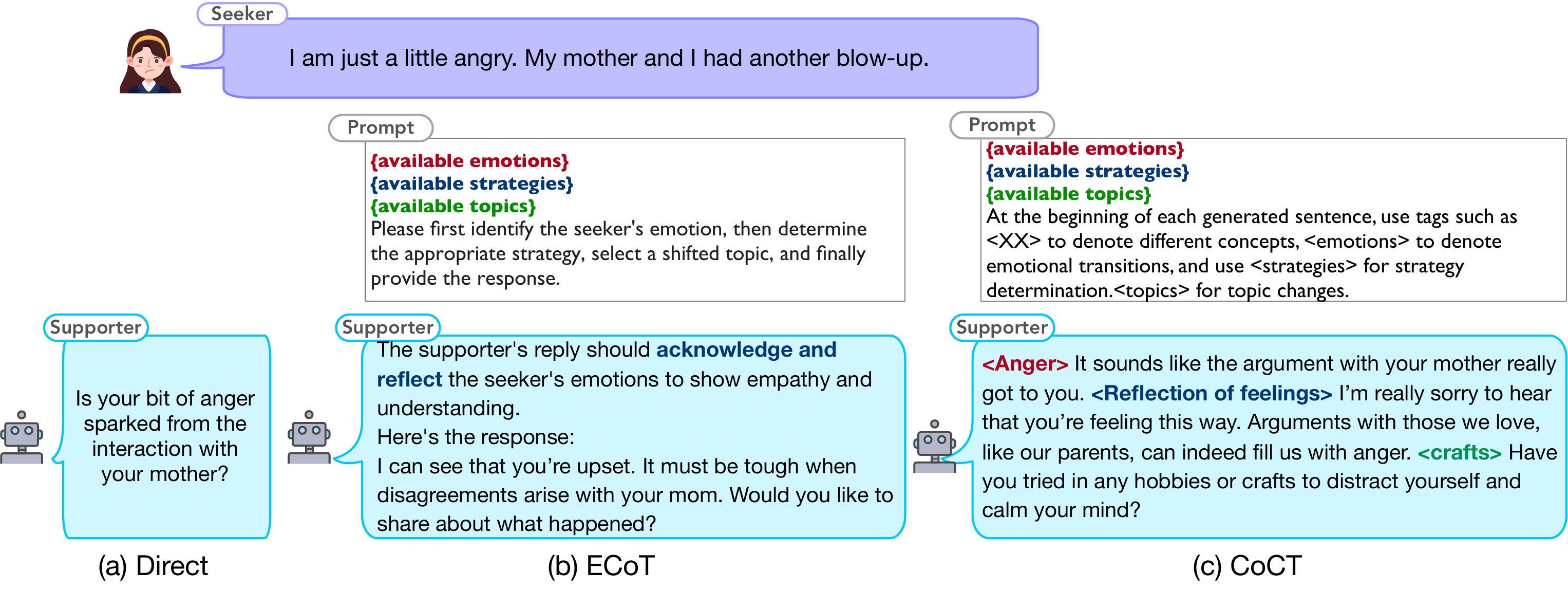} 
  \caption{The paradigm of {\ModelName}. The red bold text represents emotions, the blue bold text represents strategies, and the green bold text represents topics of conversation transitions.}
  \label{fig:paradigm}
\end{figure*}

To bridge this gap, we aim to propose an alternative, generalized prompting paradigm that enhances LLMs in broader, open domains, such as daily and emotional support conversations. Our idea is motivated by the hierarchical nature of human thinking, where a person generally first formulates a list (or chain) of strategic concepts, each of which may then guide detailed thoughts \cite{lcmteam2024largeconceptmodelslanguage,ning2024skeletonofthought}. This chain of concepts and corresponding concept-guided contents formulates the ultimate human thought. For example, a customer may deliberately first exaggerate the product's disadvantages, then make a solid decision during a bargain; a psychological counselor may first express an empathetic attitude, then provide a detailed, practical solution. In daily life, we summarize these concepts into several primary types, including:
%In Open-ended conversations, such concepts may consist of 
\begin{itemize}
    \item \textbf{Emotion}: the agent's emotional or sentiment states, such as joy, sadness, or anger. 
    \item \textbf{Strategy}: the agent's conversational strategy, such as asking for more details or summarizing the contents.
    \item \textbf{Topic}: the agent might choose to deepen the topic or shift the user's intention to a new topic for better effects.
\end{itemize}
Table \ref{tab:dataset_concepts} exemplifies these concept terms from the annotations of previous textual datasets \cite{liu2021ESConv,li-etal-2017-dailydialog,rashkin-etal-2019-towards,zhou-etal-2024-think}. We argue that LLM can utilize these concept annotations in the new prompt paradigm, resulting in a new hierarchical way of thinking.

%For a chain of thought flow, it might first be the concept transitions that leads to the transition of detailed expressions. In daily life, such concepts could be strategies, plans, or emotions, determined by the detailed scenario. 

In this paper, we propose a straightforward but effective paradigm called \textbf{C}hain \textbf{o}f \textbf{C}onceptual \textbf{T}hought (\textbf{{\ModelName}}), which introduces a \textit{conceptual chained} variant of the classical CoT, as the solution to open-domain or daily conversations. For each response utterance, we prompt the LLM to first propose a concept tag (might be one instance of either emotion, strategy, or topic, surrounded by special tokens), then generate the content grounded by that concept. Figure \ref{fig:result_snapshot} provides a quick case comparison between Direct inference, ECoT and {\ModelName}. {\ModelName} showcases in both open-domain and emotional support conversation tasks, either given a pre-defined, fixed concept list (refer to Table \ref{tab:dataset_concepts}), or based on dynamic concept generation. {\ModelName} also generalizes well to out-of-domain concepts and queries. Based on automatic, human and LLM-based evaluations, {\ModelName} outperforms other prompt-based baselines, including Direct inference, Self-Refine, ECoT, SoT and RAG. Figure \ref{fig:result_snapshot} provides a quick snapshot of the human scoring comparison. To summarize, our key contributions include: 
%  Figure \ref{fig:result_snapshot} provides a quick case comparison between direct inference, ECoT and {\ModelName}. {\ModelName} showcases in open-domain conversation datasets, either given a fixed concept list (from Table \ref{tab:dataset_concepts}), or by dynamic concept generation.
\begin{itemize}
    \item We propose {\ModelName}, a prompt-based paradigm which allows the LLM to think conceptually and strategically in daily conversation or open-domain tasks.
    \item We test {\ModelName} on tasks of daily conversation and emotional support conversations, and obtain better response quality than strong baselines.
    %\item Cases further indicate that {\ModelName} has a more human-like way of thinking, guiding the self-emotion or self-plan transitions.
    \item{\ModelName} is robust to out-of-domain concepts and queries, and can also perform reasonably with concepts generated on-the-fly. 
\end{itemize}

\section{Method}
\label{sec:method}
This section introduces our method details, including the prompt paradigm, the possible concepts, and resulting inner-utterance concept transitions. 

%Figure \ref{fig:paradigm} shows an example of {\ModelName}, compared to the direct inference and CoT results. 

\subsection{Chain-of-Conceptual-Thought Prompt}

According to the studied scenario, we assume a pre-defined list of concepts $\{ c \}$ is available based on prior knowledge. These concepts are then applied to steer the LLM generation
\begin{equation}
    output \leftarrow \text{LLM} (\{c_1, c_2, \cdots, c_n \}, \mathcal{I}, input)
\end{equation}
in which $\mathcal{I}$ is the instruction template of {\ModelName}, $c$ is a concept and $n$ is the total number of concepts. 

Below is the template of our {\ModelName} prompt:

\begin{tcolorbox}[title=CoCT Prompt Format,
    colback=white,
    colframe=yellow!75!black,
    colbacktitle=yellow,
    coltitle=black,
    breakable,
    label=tc:vanilla_prompt,
    fonttitle=\bfseries]
    \hspace*{1em}\texttt{`\{possible concepts\}',} \\
    \hspace*{1em}At the beginning of each generated response, use tags such as <XX> to denote different concepts, then follow the sentence content.\\
    \hspace*{1em}Use chain of concepts to denote the concept transitions.
\end{tcolorbox}

To summarize, the {\ModelName} prompt $\mathcal{I}$ starts from a detailed definitions of the conceptual list $\{ c \}$, which helps the LLM extract usable concepts from potentially unlimited space. After that, for each generated sentence, we ask the LLM to first decode the special token of concept, then decode the detailed sentence content. 

\subsection{Formulation of Concepts} 

As discussed in Section \ref{sec:introduction}, in this paper we study three main concept types: \textbf{emotion}, \textbf{strategy} and \textbf{topic}. While we find {\ModelName} can also work with concepts generated on-the-fly, better performance can be obtained with \textit{pre-defined} concept lists. As indicated by Table \ref{tab:dataset_concepts}, we get the concepts of \textbf{emotion} and \textbf{strategy} directly from the annotations of ESConv and DailyDialogue. For \textbf{topic}, we refer to the topic guidance study in Cskills, which provides about 8 topics after frequency filtering: \textit{sports, travel, art, music, technology, food and drink, hobbies and crafts, entertainment} and \textit{animal}.  

%Conversation topics may be divergent and open-ended. To obtain a fixed list,  

%EmpatheticDialogues lacks the strategy annotation; as a result, ESconv and DailyDialogue can be employed as in-domain (ID) evaluation of strategy, while EmpatheticDialogues is an out-of-domain (OOD) evaluation. %we use the emotion and strategy lists of ESconv when testing on EmpatheticDialogues. Such configurations employ ESconv and DailyDialogue as in-domain (ID) evaluation, and EmpatheticDialogues as out-of-domain (OOD) evaluation.

\subsection{{\ModelName} Elicits the Inner-utterance Transitions} 

\begin{figure}[!t]
\centering
  \includegraphics[width=0.98\linewidth]{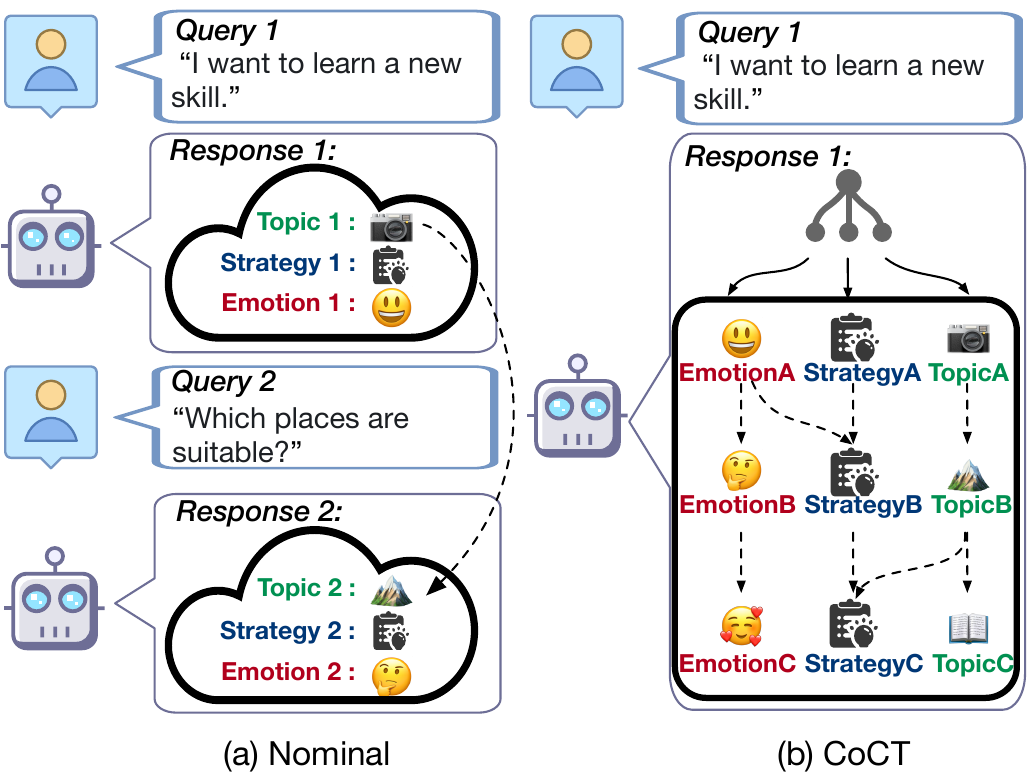} 
  \caption {Comparative visualizations between outer-utterance transitions (Left) and inner-utterance transitions (Right). The dash arrows indicate the conceptual transitions between emotions, strategies or topics.}
  \label{fig:inner_utterance_transition}
\end{figure}

Human talk sometimes implies the transition of concepts within a single utterance; however, nominal LLM-based conversation usually fails to do so. Instead, one can observe that each of their responses contains a single emotion or topic, and such concepts' transitions happen on the outer-utterance (Figure \ref{fig:inner_utterance_transition} (Left)). On the other hand, {\ModelName} naturally encourages the inner-utterance concept transitions (Figure \ref{fig:inner_utterance_transition} (Right)), which is part of our main purpose:
\begin{equation}
    c_i \rightarrow c_j, \quad i, j \in \{1, \cdots, n\}
\end{equation}

\section{Experiment}
\label{sec:experiment}

In this section, we first briefly introduce the experimental setting, including the implementation, baseline and metric details. Then we exhibit the effect of {\ModelName} on daily conversations. To provide more quantitative results, we then conduct more experiments on emotional support conversations. Finally, we provide some insightful analysis.

%Further details are provided in the Appendix.

\subsection{Setting}

\paragraph{Implementation.} The experiment is conducted on vLLM, with 8 A100 80GB PCIe GPUs. The context window is restricted to 4096. To validate the effectiveness of {\ModelName} on different model scales, we experiment with the performances on Llama3-8B-instruct \cite{llama3modelcard}, Llama3-70B-instruct \cite{llama3modelcard} and GPT4o, respectively.

\paragraph{Datasets and tasks.} In this paper, we conduct {\ModelName} on both in-domain (ID) and out-of-domain (OOD) tasks. In ID tasks, the datasets have annotations of emotions, strategies, or topics, which are used as the possible concepts $\{ c \}$ during the test. For OOD tasks, the datasets do not have such concept annotations. Therefore, we use $\{ c \}$ defined in ID tasks, and test {\ModelName}'s generalization capability of these concepts to this new dataset. Since it is more common to assume there are no conceptual annotations, the OOD tasks can reflect the capability of {\ModelName} in most scenarios. The conversational datasets we use can be classified into two main categories:

%\textbf{\textcolor{red}{add intro of datasets @guqingqing(done)}} We evaluate the performance of our method on two main categories of dialogue datasets, covering both in-domain (ID) and out-of-domain (OOD) scenarios to assess generalization ability. The datasets are selected to span Open-ended conversation and emotional support dialogue tasks, ensuring comprehensive validation of model behavior across different interaction contexts.  

%\begin{itemize}
%    \item 

\noindent (1) Open-ended conversations: we use the annotated lists of emotions and strategies from \textit{DailyDialogue} \cite{li-etal-2017-dailydialog}, and the list of topics from \textit{Cskills} \cite{zhou-etal-2024-think}. Together, we have 7 emotions, 4 strategies and about 20 topics. Correspondingly, we conduct the ID test of {\ModelName} on the test set of \textit{DailyDialogue} and \textit{Cskills}. After that, we apply {\ModelName} on another open-topic dataset \textit{MultiWoZ} \cite{budzianowski-etal-2018-multiwoz} as the OOD test, which does not have concept annotations.
    %including \textit{DailyDialogue} \cite{li-etal-2017-dailydialog} (ID) which focuses on daily topics and is annotated with 7 emotions and 4 strategies; the Cskills benchmark \cite{zhou-etal-2024-think} (ID) with about 1000 queries and around 20 topics; as well as \textit{MultiWoZ} \cite{budzianowski-etal-2018-multiwoz} (OOD), with topics such as travel and entertainment, and without concept annotations.

    %\item 

\noindent (2) Emotional support conversations: we then investigate the performance of {\ModelName} on this specific scenario for more in-depth analysis. We employ 11 emotions and 8 strategies from \textit{ESConv}, conduct the ID test of {\ModelName} on the test of \textit{ESConv}; and the OOD test on \textit{EmpatheticDialogues} \cite{rashkin-etal-2019-towards}.
    %we employ \textit{ESConv} \cite{liu2021ESconv} as ID dataset, with annotation of 11 emotions and 8 strategies; and \textit{EmpatheticDialogues} \cite{rashkin-etal-2019-towards} as OOD dataset.
%\end{itemize}

Appendix \ref{appendix:dataset} provides more dataset details.

\paragraph{Metrics.} Besides the cases, the response quality can also be evaluated by human annotations, automatic metrics and LLM. Here we briefly introduce the details.
%Besides the cases, the response quality can also be evaluated by human annotations, automatic metrics and LLM. Here we briefly introduce the details.

\begin{itemize}
    \item Automatic Metrics: Include similarity-based metrics such as BLEU-2 (\textbf{B-2}) \cite{papineni2002bleu}, Rouge-L (\textbf{R-L}) \cite{lin2004rouge} and CIDEr (\textbf{CDr}) \cite{vedantam2015cider}; and diversity-based metric Distinct-2 (\textbf{D-2}).
    %to verify if our response could be close to the `golden' human annotation, given suitable conceptual guidance, and also evaluate the response quality directly by LLM simulation, LLM-as-a-Judge, and human annotations.

    %BLEU-2 \cite{papineni2002bleu} first computes the geometric average of the modified $n$-gram precisions, then calculates the BP and BLEU-2 result. Rouge-L \cite{lin2004rouge} uses LCS-based F-measure to estimate the similarity between two summaries. CIDEr \cite{vedantam2015cider} calculates the cosine similarity from the average of different $n$-grams.

    \item LLM-Based Simulations: To qualitatively evaluate the quality of daily conversation, we utilize the Cskills benchmark \cite{zhou-etal-2024-think}, with about 1000 queries. The agent is expected to generate reasonable and qualified dialogue content. Simulation stops when there is no significant intention to communicate. The averaged utterance length and dialogue rounds are employed as metrics.
    % expects the agent to generate reasonable and qualified dialogue content.
    %To qualitatively evaluate the quality of Open-ended conversation, we utilize the Cskills benchmark \cite{zhou-etal-2024-think}, with about 1000 queries. The agent is expected to generate reasonable and qualified dialogue content. Simulation stops when there is no significant intention to communicate. The averaged utterance length and dialogue rounds are employed as metrics.

    \item LLM-as-a-Judge: We used GPT-4o to compare different responses and provide the win-tie-lose suggestion. The evaluation prompt is from \cite{madani2024steeringconversationallargelanguage}.
    
    \item Human Scoring: four annotators are employed to provide quality scoring. The cross-validation is conducted with the averaged results shown later. Similar to \cite{kang-etal-2024-large}, the scoring dimensions include \textit{Acceptance}, \textit{Effectiveness}, \textit{Sensitivity}, \textit{Fluency}, and \textit{Emotion}, and the \textit{user's Satisfaction}. %Scores are averaged over ESConv, and EmpatheticDialogues.
\end{itemize}

We provide more evaluation details in Appendix \ref{appendix:evaulation}, such as computation formula, annotation principle and evaluation prompts.

\paragraph{Baselines.} As fair comparisons, we implement several prompt-based baselines, with the same base model and context. Such baselines can be classified into several categories:
(1) \textit{Direct} inference the LLM, \textit{Direct-Refine} the result within the same response, and Self-Refine \cite{Madaan2023SelfRefine} the result by inference the second time.
(2) Methods based on chained or other structured data format, including \textit{ECoT} \cite{li2024enhancingemotionalgenerationcapability}, \textit{SoT} \cite{ning2024skeletonofthought}, \textit{ToT}  \cite{yao2023tree}, and \textit{Plan-and-Solve} \cite{wang2023planandsolvepromptingimprovingzeroshot}.
(3) Retrieval augmented generation (\textit{RAG}) \cite{fan2024RAGmeetingLLMs} which retrieves the suitable concepts and descriptions and prompts the LLM to respond. In this work, we employ E5-Large \cite{wang2024E5} as the retriever.
(4) \textit{CSIM} \cite{zhou-etal-2024-think}: a self-chat method which produces the response based on its implicit simulation of future dialogue. 

More details can be found in Appendix \ref{appendix:baseline}.

\begin{table*}[t!]
\caption{Results on automatic metrics on daily conversations (DailyDialog and MultiWoZ) and ESC (ESConv and EmpatheticDialogues). E.D. abbreviates EmpatheticDialogues. The best results on each LLM are \textbf{bolded} and the second best are \underline{underlined}. % $\mathcal{Q}$, $\mathcal{B}$ A single strategy planner is employed to predict strategies and provides them to each LLM. $^{\ast}$: results are obtained directly from \cite{kang-etal-2024-large}.
%\textbf{\textcolor{red}{complete the values; add new baselines; add MultiWoZ in this table (after DailyDialogue) @guqingqing(done)}} 
}
\label{tab:quantitative_results}
\renewcommand{\arraystretch}{1.11} 
\centering
% \small
\resizebox{1.99\columnwidth}{!}{
\begin{tabular}{l | ccc | cccc | ccc | cccc } 
    \toprule
    \multirow{2}{*}{Method $\downarrow$} &  
    \multicolumn{3}{c|}{DailyDialogue (ID)} &
    \multicolumn{4}{c|}{MultiWoZ (OOD)} & 
    \multicolumn{3}{c|}{ESConv (ID)} &  
    \multicolumn{4}{c}{EmpatheticDialogues (OOD)} \\ 
    \cmidrule{2-4} \cmidrule{5-8} \cmidrule{9-11} \cmidrule{12-15}  
    & B-2 & R-L & CDr & B-2 & R-L & D-2 & CDr & B-2 & R-L & CDr & B-2 & R-L & D-2 & CDr  \\ 
    %& original & N/A & 1.151  & N/A & N/A  \\
    \midrule   %LLaMA3-8B-Instruct
   \multicolumn{15}{l}{\textit{LLaMA3-8B-Instruct}} \\  % t (\textit{0-shot})
    %\midrule   
    \;+ Direct          & \underline{3.35} & \underline{10.33} & 1.99 & \underline{4.94} & \underline{13.18} & 16.09 & 4.12 & \underline{3.47} & \underline{10.64} & 1.14 & \underline{3.09} & \underline{9.91} & 25.23 & \bf 1.60 \\ 
    \;+ Direct-Refine   & 2.56 & 8.70 & 1.75 & 4.02 & 11.90 & 17.92 & 1.57 & 3.10 & 6.13 & 1.30 & 2.56 & 9.12 & 22.32 & 0.42 \\ 
    \;+ Self-Refine     & 2.40 & 7.75 & 1.27  & 4.47 & 12.61 & 19.13 & 3.07 & 3.34 & 9.71 & \underline{1.79} & 3.08 & \underline{9.91} & 25.20 & \underline{1.56} \\ 
    \;+ ECoT            & 1.78 & 6.00 & \underline{2.43} & 4.22 & 11.89 & \bf 26.28 & 2.36 & 3.16 & 10.50 & 1.50 & 2.91 & 9.79 & \underline{32.65} & 1.37 \\ 
    \;+ Plan-and-Solve             & 2.60 & 7.76 & 1.92 & 4.32 & 10.74 & 18.99 & \underline{4.87} & 2.81 & 8.27 & 1.59  & 2.69 & 6.93 & 24.02 & 0.99 \\ 
    \;+ ToT             & 2.52 & 8.84 & 0.61 & 3.06 & 10.05 & 15.98 & 0.23 & 2.65 & 9.81 & 0.60 & 2.31 & 9.05 & 29.09 & 0.12 \\ 
    \;+ SoT             & 2.53 & 7.97 & 1.76 & 3.58 & 9.77 & 17.09 & 3.55 & 3.07 & 8.76 & 1.70 & 1.79 & 5.66 & \bf 48.59 & 0.19 \\ 
    \bf \;+ {\ModelName} (ours) & \bf 3.62 & \bf 11.64 & \bf 3.84 & \bf 6.94 & \bf 17.95 & \underline{19.79} & \bf 9.73 & \bf 4.75 & \bf 13.31  & \bf 4.87 & \bf 3.29 & \bf 10.84 & 26.27 & 1.08 \\
    \midrule
    \multicolumn{15}{l}{\textit{LLaMA3-70B-Instruct}} \\
    \;+ Direct          & 3.50 &11.20 & 2.46 & \underline{5.19} & \underline{14.38} & 14.38 & 1.37 & 3.06 & 10.16 & 0.8 & 2.60 & 8.90 & 26.71 & 0.44 \\  
    \;+ Direct-Refine   & \bf 4.16 & \bf 12.84 & \bf 6.04 & 4.63 & 13.32 & 15.65 & 1.40 & 2.54 & 9.40 & 0.27 & 2.53 & 9.07 & 25.28 & 0.30 \\
    \;+ Self-Refine     & 3.40 & 10.83 & 2.97 & 4.82 & 13.47 & 17.09 & 1.51 & 2.97 & 10.12 & 1.07 & \underline{2.81} & \underline{9.74} & 27.06 & 0.48 \\
    \;+ ECoT            & 2.09 & 7.62 & 1.23 & 4.82 & 13.47 & 17.09 & 1.51 & 1.85 & 9.67 & 1.03 & 2.60 & 9.06 & \bf 34.02 & \underline{1.14} \\
    \;+ Plan-and-Solve             & 2.13 & 6.32 & 0.80 & 3.96 & 10.41 & 13.49 & 3.09 & 2.01 & 8.68 & 1.36 & 2.14 & 6.25 & 37.16 & 0.50 \\ 
    \;+ ToT             & 2.68 & 9.43 & 0.69 & 4.08 & 13.91 & 16.62 & 0.40 & 2.39 & 8.96 & 0.16 & 2.16 & 8.24 & 27.14 &0.13 \\ 
    \;+ SoT             & 2.72 & 8.55 & 1.72 & 3.92 & 11.01 & \bf 20.80 & \underline{3.96} & 3.13 & 9.49 & 1.16 & 2.12 & 6.83 & 45.98 & 0.35 \\ 

    \;+ RAG             & 2.78 & 9.67 & 1.67  & - & - & - & - & \underline{3.44} & \underline{10.95} & \underline{1.65} & 2.10 & 8.18 & 28.92 & 0.13 \\
    \bf \;+ CoCT (ours) & \underline{3.87} & \underline{11.94} & \underline{3.81} & \bf 6.45 & \bf 17.00 & \underline{17.34} & \bf 4.78 & \bf 4.30 & \bf 12.71 & \bf 2.07 & \bf 3.48 & \bf 11.20 & \underline{29.34} & \bf 1.45\\ 
    \midrule
    \multicolumn{15}{l}{\textit{GPT-4o}}  \\
    \;+ Direct & \underline{4.25} & \underline{12.80} & 6.09 & - & - & - & - & 3.48 & \underline{13.21} & \underline{3.83} & \underline{3.24} & \underline{10.70} & \underline{36.64} & \bf 1.54  \\  %  & 4.08 & 12.09
    \;+ Direct-Refine  & 4.03 & 12.22 & 4.49 & - & - & - & - & 4.19 & 12.61 & 2.65 & - & - & - & - \\
    \;+ Self-Refine & 3.19 & 9.84 & 3.19 & - & - & - & - & \bf 5.40 & 12.70 & 2.82  & - & - & - & - \\
    \;+ ECoT & 3.48 & 11.10 & 3.46 & - & - & - & - & 3.64 & 11.64 & 1.95 & 2.30 & 7.84 & 26.87 & 0.07 \\
    \;+ RAG & 4.15 & 9.75 & \underline{7.14} & - & - & - & - & 3.68 & 11.05 & 3.67 & 2.60 & 8.63 & \bf{53.66} & 0.95 \\ %  & 2.91 & 9.79
    \bf \;+ CoCT (ours) & \bf 4.60 & \bf 13.25 & \bf 8.73 & - & - & - & - & \underline{4.62} & \bf 13.77 & \bf 4.79 & \bf 3.47 & \bf 11.20 & 32.36 & \underline{1.30} \\
    \bottomrule
\end{tabular}
}

\end{table*}

\subsection{Results on Daily Conversation}

We first test {\ModelName}'s performance on daily conversations, with no explicit constraints on topics. 
%, with no explicit constraints on topics. 

%We start from an annotated dataset, DailyDialogue \cite{li-etal-2017-dailydialog}, which provides daily topics with emotion and strategy (named `act' in the original dataset) annotated along with the response. We employed all its emotions and strategies as the concept lists $\{ c \}$, with detailed values exhibited in Table \ref{tab:dataset_concepts}. 

\paragraph{Performance on in-domain concepts.}

Table \ref{tab:quantitative_results} shows the automatic metrics of DailyDialogue, where {\ModelName} performs the best with the basis on LLaMA3-8B-Instruct and GPT-4o, indicating it can generate responses in a similar style to the original annotated response. {\ModelName} performs the second-best on LLaMA3-70B-Instruct (worse than Direct-Refine), which we suppose may be due to insufficient exploration when we apply the prompt tested on 8B directly to 70B.
%the strong capability of LLaMA3-70B-Instruct for long context generation, while combining {\ModelName} with the refine mechanism would produce even higher performance. 

\paragraph{Generalize with out-of-domain concepts or queries.} Table \ref{tab:quantitative_results} also exhibits the result of MultiWOZ on LLaMA3-8B-Instuct and LLaMA3-70B-Instuct. {\ModelName} still performs the best on B-2, R-L and CDr, indicating its strong generalization capabilities on different topics and domains. On the other hand, {\ModelName} is the second-best on D-2, a diversity-based metric, which is reasonable since the pre-defined concepts may constrain the methodology's scope\footnote{We will discuss the improvement method in Section \ref{sec:discussion}.}.

%\cite{budzianowski-etal-2018-multiwoz} is a human-human written conversation dataset spanning over multiple domains and topics, but without any annotation of concepts. We test CoCT on it with emotion and strategy lists from DailyDialogue, and the topic list from Cskills (see Table \ref{tab:dataset_concepts}). Table \ref{tab:multiwoz_case} shows a typical case of {\ModelName} on the MultiWOZ dataset, which indicates that {\ModelName} still performs well with these OOD concepts.

\begin{table}[t!]
\caption{Results on the Cskills. We use the `concept guidance' subset of the original benchmark.}
% Results of CSIM are obtained directly from \cite{zhou-etal-2024-think}.
\label{tab:cskills_result}
\centering
\small
%\resizebox{0.98\columnwidth}{!}{
%\begin{tabular}{c | cc }
%    \toprule
%    Method & AvgLen & Rounds \\ %
%    \midrule
%    Direct & 46.41 &	3.53  \\
%    CSIM &	47.73 &	4.26 \\ 
%    RAG & 49.13 & 4.36 \\
%    \textbf{\ModelName} (ours) & \bf67.04 & \bf	4.62 \\
%    \bottomrule
%\end{tabular}

\begin{tabular}{c | cccc }
    \toprule
    Method $\rightarrow$ & Direct & CSIM & RAG & \textbf{\ModelName} (ours) \\ %
    \midrule
    AvgLen & 46.41 & 47.73 & 49.13 & \bf 67.04	  \\
    Rounds & 3.53 &	4.26 & 4.36 & \bf 4.62 \\
    \bottomrule
\end{tabular}
%}
\end{table}

\begin{table*}[h!]
    \caption{Open multi-turn cases generated by {\ModelName}.} 
    \label{tab:case}
    \centering
    \small
    %\resizebox{\textwidth}{!}{%
    \begin{tabular}{c|c|l}
        \Xhline{2\arrayrulewidth}
        %\multicolumn{2}{c}{\textbf{w/ strategy}}  \\ 
        %\hline
        \multicolumn{1}{c|}{\multirow{9}[2]{*}{\rotatebox[origin=c]{90}{\centering w/ strategy}}} 
        & user & \makecell[l]{\textit{I've started learning how to cook.}} \\
        \cmidrule{2-3}
        & {\ModelName} & \makecell[l]{\textbf{<Affirmation and Reassurance>} That's fantastic! Learning to cook is\\ such a valuable skill and can be really rewarding. \\ \textbf{<Question>} What type of dishes are you interested in trying out?}   \\
        \cmidrule{2-3}
        & user & \makecell[l]{\textit{I tried making pizza last week. It was a bit of a fail the first time, but it still tasted okay!}}    \\ 
        \cmidrule{2-3}
        & {\ModelName} & \makecell[l]{\textbf{<Reflection of feelings>} It sounds like you were a bit disappointed with the first\\ attempt, but also glad that it still turned out tasty. \\ \textcolor{blue}{\textbf{<Humor>}} Hey, even a "fail" pizza is still pizza, right? That's a win in my book! \\ \textbf{<Question>} What part of making the pizza did you find the most challenging?}   \\
        %\hline
        %seeker & \makecell[l]{\textit{I put too much sauce on it, so it ended up a little salty.}}  \\
        %\hline
        %supporter & \makecell[l]{\textbf{<Restatement>} So, it sounds like the pizza ended up a bit too salty because of the extra sauce. \\ \textbf{<Providing Suggestions and Information>} A tip for next time might be to use a bit less sauce and spread it more evenly. \\ \textbf{<Affirmation and Reassurance>} But remember, it's all part of the learning process! \\ Every little mistake is a step towards becoming a better cook. What do you plan to try next?}  \\

        \Xhline{2\arrayrulewidth}
        %\multicolumn{2}{c}{\textbf{w/ emotion}}  \\ 
        %\hline
        \multicolumn{1}{c|}{\multirow{7}[2]{*}{\rotatebox[origin=c]{90}{\centering w/ emotion}}} 
        & user & \makecell[l]{\textit{I've started learning how to cook.}} \\
        \cmidrule{2-3}
        & {\ModelName} & \makecell[l]{\textbf{<Excitement>} That's fantastic! Learning to cook is such a rewarding skill.\\ What dish are you most excited to try making?}   \\
        \cmidrule{2-3}
        & user & \makecell[l]{\textit{I tried making pizza last week. It was a bit of a fail the first time, but it still tasted okay!}}   \\
        \cmidrule{2-3}
        & {\ModelName} & \makecell[l]{\textbf{<Nervousness>} First attempts can be tricky, but it's great that it still tasted okay! \\ \textbf{<Encouragement>} Practice makes perfect, and every time you cook, you'll get better. \\ What part of making the pizza did you find most challenging?}   \\
        %\hline
        %seeker & \makecell[l]{\textit{I put too much sauce on it, so it ended up a little salty.}}   \\
        %\hline
        %supporter & \makecell[l]{\textbf{<Sadness>} That sounds a bit disappointing, but \\ \textbf{<Encouragement>} it's a common mistake, and now you know for next time! \\ \textbf{<Support>} Adjusting the amount of sauce can make a big difference. What kind of toppings did you use?}   \\
        \Xhline{2\arrayrulewidth}
        %\multicolumn{2}{c}{\textbf{w/ emotion+strategy}}  \\ 
        %\hline
        \multicolumn{1}{c|}{\multirow{8}[2]{*}{\rotatebox[origin=c]{90}{\centering w/ emotion+strategy}}} 
        & user & \makecell[l]{\textit{I've started learning how to cook.}}  \\
        \cmidrule{2-3}
        & {\ModelName} & \makecell[l]{\textcolor{blue}{<\textbf{Interest}}> That's fantastic! Cooking can be a rewarding and enjoyable skill to develop. \\ \textbf{<Questioning>} What kinds of dishes are you learning to make?}   \\% such
        \cmidrule{2-3}
        & user & \makecell[l]{\textit{I tried making pizza last week. It was a bit of a fail the first time, but it still tasted okay!}}    \\ 
        \cmidrule{2-3}
        & {\ModelName} & \makecell[l]{\textbf{<Affirmation>} It's great that you tried making pizza! \\ \textbf{<Reassurance>} Everyone has a few mishaps when they're learning something new. \\ \textcolor{blue}{\textbf{<Humor>}} Besides, an "okay" tasting pizza is still pizza, right? \\ \textbf{<Questioning>} What part of the process do you think was the most challenging?}   \\
        %\hline
        %seeker & \makecell[l]{\textit{I put too much sauce on it, so it ended up a little salty.}}  \\
        %\hline
        %supporter & \makecell[l]{<Reflection of feelings> I can understand how that might have been disappointing, \\ especially when you were looking forward to enjoying your homemade pizza. \\ <Self - disclosure> I once did the same thing with a pasta dish - it was so salty! \\ <Providing Suggestions and Information> Next time, you might try adding less sauce and then taste-testing as you go. \\ Cooking is full of little adjustments and learning experiences. \\ <Questioning> What other recipes are you thinking about trying?}  \\
        \Xhline{2\arrayrulewidth}
        
    \end{tabular}
    %}

\end{table*}

\paragraph{Cases on more situations} To further verify the generalization capability of {\ModelName}, in this experiment we test it with arbitrary user queries. To explore more situations, we experiment with three combinations of concept types: with strategy only (w/ strategy), with emotion only (w/ emotion), or with both emotion and strategy (w/ emotion+strategy). Furthermore, we also attempt to expand the strategies with two more arbitrary tags: \textcolor{blue}{<Humor>} and \textcolor{blue}{<Interest>}. We expect such new tags can be automatically recognized by LLM and provide more interesting responses.

Table \ref{tab:case} showcases {\ModelName}'s performance on these arbitrary scenarios. It can be observed that {\ModelName} still performs well on these out-of-domain queries and concepts, generating reasonable concept transitions. Especially, <Humor> is utilized in the first case and <Interest> is utilized in the third case, generating more funny responses and steering the conversation to a higher communication level. %Also, {\ModelName} is robust with different choices of concept types, performing well in all three cases. 

%qualitatively evaluate the dialogue quality, we also utilize the Cskills benchmark \cite{zhou-etal-2024-think}. The agent is expected to generate reasonable and qualified dialogue contents, and stops when there is no significant intention to communicate. The averaged utterance length and dialogue rounds are employed as metrics. We employ GPT4o as the simulator and compare to the original method (CSIM) in \cite{zhou-etal-2024-think}.

%Table: finetuned result

%Table \ref{tab:prompt_comparison_score}.

%By combining different responses with history appropriately, we made GPT-4o select more appropriate responses. %We refer to the prompt of 

%\cite{madani2024steeringconversationallargelanguage}, see Table \ref{tab:prompt_comparison_score} for details.

\subsection{Results of Emotional Support Conversations}

%In order to provide more quantitative evaluations, 

In this subsection, we further investigate the capability of {\ModelName} on a specific domain of conversations, the emotional support conversation (ESC). %We employ two classical datasets, ESConv and EmpatheticDialogues.

%\paragraph{Datasets and Benchmarks.} 

%Figure: model size VS score (wrt base model type), Standard prompting VS COT VS COCT VS fine-tuning best
%finetune llama3 with GPT-4 labeled response

\paragraph{Automatic Metrics.} The last two columns of Table \ref{tab:quantitative_results} illustrate the ID results on ESconv and the OOD results on EmpatheticDialogues. Similar to tests on daily conversations, {\ModelName} is still the best on B-2, R-L and CDr. This observation indicates {\ModelName} holds stable and robust performance for this specific domain.

%with all base models. {\ModelName} generally performs the best on B-2, R-L and CDr. After that, we conduct the OOD experiment of EmpatheticDialogues on LLaMA3-70B-Instruct, as shown by the last column. {\ModelName} still outperforms other baselines on B-2, R-L and CDr, and performs the second best on the diversity-based metric (D-2). 

%\paragraph{LLM-as-a-Judge.} 

\begin{table*}[htbp!]
\caption{Human evaluation of response quality on ESConv and EmpatheticDialogues. The best results of each LLMs are \textbf{bolded} and the second best are \underline{underlined}.}
\label{tab:response_quaility}
%${\ast}$: Value with bold indicates the best results while value with underline indicates the second best results. We do not assess non-LLM methods because they are not open-sourced.
\centering
\resizebox{\textwidth}{!}{%
\begin{tabular}{c|l|ccccccc}
    \toprule
    \multirow{2}[2]{*}{Size} & \multicolumn{1}{c|}{\multirow{2}[4]{*}{Method}} & \multicolumn{7}{c}{Human Annotation} \\
%\cmidrule{1-2}  
\cmidrule{3-9}    
 &  & Fluency  & Emotion & Acceptance & Effectiveness & Sensitivity & Alignment & Satisfaction \\ % Interesting % Total % $\uparrow$
    \toprule   
    - & orignal dataset & 3.51 & 3.61 & 3.40 & 3.10 & 3.50 & 3.20 & 3.30 \\
    \midrule
    %\multicolumn{1}{c|}{} & Prompt+Qwen2 & 72B  & - & - & - & - & - & - & - \\
    %\multicolumn{1}{c|}{} & Prompt+Mistral & 8x7B  & - & - & - & - & - & - & - \\
    \multicolumn{1}{c|}{\multirow{5}[2]{*}{8B}} 
    & LLaMA3 (8B)  & 2.95 & 3.00 & 2.60 & 2.40 & 2.70 & 2.70 & 2.60 \\
    & \;+ Direct-Refine & 3.09 & 3.09 & 2.73 & 2.91 & \underline{2.91} & 2.82 & \underline{2.84} \\
    & \;+ Self-Refine & \bf 3.10 & \underline{3.15} & 2.80 & 2.70 & 2.90 & 2.80 & 2.80 \\
    & \;+ ECoT  & \underline{3.08} & 3.08 & \underline{2.83} & 2.67 & \underline{3.00} & \underline{2.83} & 2.83 \\
    %& \;+ {\vanillaFT} & 3.2 & 3.4 & 2.7 & 2.7 & 2.9 & 3.3 & 2.9 \\
    %& \;+ CoT+SFT  & 3.7 & 3.6 & 3.2 & 3.7 & 3.6 & 3.4 & 3.5 \\
    %& \;+ {\ModelName}-emo & 3.1 & 3.4 & 2.6 & 2.6 & 3.3 & 2.7 & 2.8 \\
    & \;+ \textbf{{\ModelName}} (ours) & \bf 3.10 & \bf 3.25 & \bf 3.00 & \bf 3.20 & \bf 3.10 & \bf 3.00 & \bf 3.08 \\ % -strategy
    %& \;+ {\ModelName}-emo+strategy & 3 & 2.9 & 3.1 & 3.1 & 2.9 & 3 & 3 \\
    \midrule
    \multicolumn{1}{c|}{\multirow{5}[2]{*}{70B}} 
    & LLaMA3 (70B)  & 3.05 & 3.30 & 2.70 & 3.00 & 3.20 & 3.10 & 3.00 \\
    & \;+ Direct-Refine & \bf 3.40 & \underline{3.50} & \underline{3.00} & \underline{3.10} & \underline{3.10} & 2.90 & 3.03 \\
    & \;+ Self-Refine & \bf 3.40 & 3.35 & 2.90 & 3.10 & 3.00 & \bf 3.20 & \underline{3.05} \\
    & \;+ ECoT  & 3.25 & 3.30 & 2.80 & 3.00 & 3.10 & \bf 3.20 & 3.03 \\
    %& \;+ {\vanillaFT} & - & - & - & - & - & - & - \\
    %& \;+ CoT+SFT  & - & - & - & - & - & - & - \\
    %& \;+ {\ModelName}-emo & 3.25 & 4.05 & 3.25 & 6.5 & 3.6 & 3.35 & 3.35 \\
    & \;+ \textbf{{\ModelName}} (ours) & \underline{3.32} & \bf 3.77 & \bf 3.36 & \bf 3.55 & \bf 3.55 & \underline{3.18} & \bf 3.41 \\ % -strategy
    %& \;+ {\ModelName}-emo+strategy & 3.318 & 3.772 & 3.364 & 3.545 & 3.545 & 3.182 & 3.409 \\
    \midrule
    \multicolumn{1}{c|}{\multirow{5}[2]{*}{larger}}
    %& GPT-4  & 3.5 & 3.8 & 3.4 & 3.3 & 3.5 & 3.5 & 3.4 \\
    %& \;+ \textbf{{\ModelName}} (ours)  & 3.8 & 3.9 & 3.7 & 3.6 & 3.8 & 3.6 & 3.7 \\
    & GPT-4o & 3.56 & 3.67 & \bf 3.67 & 3.56 & 3.56 & 3.22 & 3.50 \\
    & \;+ Direct-Refine & \underline{3.80} & 3.70 & \underline{3.60} & 3.50 & \bf 3.70 & \underline{3.60} & \underline{3.60} \\
    & \;+ Self-Refine & \underline{3.80} & \underline{3.75} & 3.50 & \underline{3.70} & \underline{3.60} & \underline{3.60} & \underline{3.60} \\
    & \;+ ECoT & 3.75 & 3.70 & 3.50 & 3.50 & \underline{3.60} & 3.50 & 3.53 \\
    & \;+ \textbf{{\ModelName}} (ours)  & \bf 3.85 & \bf 4.00 & \underline{3.60} & \bf 4.10 & \bf 3.70 & \bf 3.70 & \bf 3.82 \\
    \bottomrule
    \end{tabular}%
}

\end{table*}

\paragraph{Human Scores.}  Table \ref{tab:response_quaility} shows the human annotation results. {\ModelName} has higher scores than direct inference, direct-refine, self-refine and CoT, with the bases of Llama3-8B-Instruct, Llama3-70B-Instruct and GPT-4. Figure \ref{fig:result_snapshot} in Section \ref{sec:introduction} provides a snapshot of the Satisfaction scores.
It is also worth mentioning that although the original dataset is originally annotated as expert demonstrations, {\ModelName} based on Llama3-70B-Instruct starts to surpass the original dataset, while other baselines can not. For results based on GPT-4o, the human scores are all higher than the original dataset, due to the stronger model basis. Our {\ModelName} still remains the best.

%\subsection{Typical Cases}

\paragraph{Ablation Study.} We conduct several ablations, including \textit{wo/ strategy} and \textit{wo/ emotion} in the concept list. We compare their performances to the formal {\ModelName} on human and automatic evaluations, as well as the pairwise evaluation (versus original dataset) by GPT-4o. Table \ref{tab:ablation_results} shows that {\ModelName} still outperforms these ablations, indicating both strategy and emotion are helpful concept types.

%To further study the functionality of the concept, we conduct another analysis where we ablate the original components of concepts in ESconv experiment, \textit{emotion} and \textit{strategy}. From the comparison evaluation versus SFT in Table \ref{tab:ablation_results}, we can observe that the full form of {\ModelName}, \textit{i.e.}, w/ emo+strategy, has higher win rate than only w/ emotion or w/ strategy. This observation suggests the implicit benefit of conceptual thinking for LLM.

\begin{table}[t!]
\caption{Ablation study of {\ModelName} on Llama3-8B-Instruct, with human, automatic and win-tie-lose rates (\%) of GPT-4o evaluation results. Experiments are conducted on ESConv. `Satis.' abbreviates Satisfaction. The best result is \textbf{bolded} and the second best is \underline{underlined}.}
% , all of which are fine-tuned on LLama3-8B-Instruct
%($\dagger$/$\ddagger$: p-value < 0.1/0.05 )}
%\textbf{\textcolor{red}{update the free concept values @guqingqing(done)}} 
\label{tab:ablation_results}
\centering
\small
\begin{tabular}{l | c | cc | cc}
\toprule
%\multicolumn{1}{l}{\textbf{}} & \multicolumn{3}{c}{\textbf{vs {\vanillaFT}}} \\ 
%\midrule
\multicolumn{1}{c|}{\multirow{2}[1]{*}{Method $\downarrow$}} & \multicolumn{1}{c|}{\multirow{1}[2]{*}{human}}   & \multicolumn{2}{c|}{\multirow{1}[2]{*}{automatic}}    & \multicolumn{2}{c}{\multirow{1}[2]{*}{VS dataset}} \\ % {\vanillaFT} 
\cmidrule{2-2} \cmidrule{3-4}  \cmidrule{5-6}
& Satis. & B-2 & R-L & win $\uparrow$   &  lose $\downarrow$     \\  %   & tie   
\midrule
%\textbf{Gold}     & 65.3\%       & 2.2\%              & 32.6\%          \\
wo/ strategy   &   2.80 & 4.25 & 12.67  & 51.90                & 37.40           \\ % & 10.7 
%\textbf{Firefly}      & 59.8\%       & 11.8\%             & 28.4\%         \\
wo/ emotion   &   3.00 & 4.03 & 12.60     & 60.90          & 33.70            \\  %  & 5.4
free concept & 2.90 & \underline{4.46} & \underline{13.14} & 66.86 & 16.75 \\
\bf {\ModelName} & \bf 3.08 & \bf 4.60 & \bf 13.34  & \bf 71.71      & \bf 26.80       \\  %  & 11.7
\bottomrule
\end{tabular}

\end{table}

\subsection{Discussion}
\label{sec:discussion}

\begin{figure*}[htbp!]
    \centering
    \includegraphics[width=0.48\linewidth]{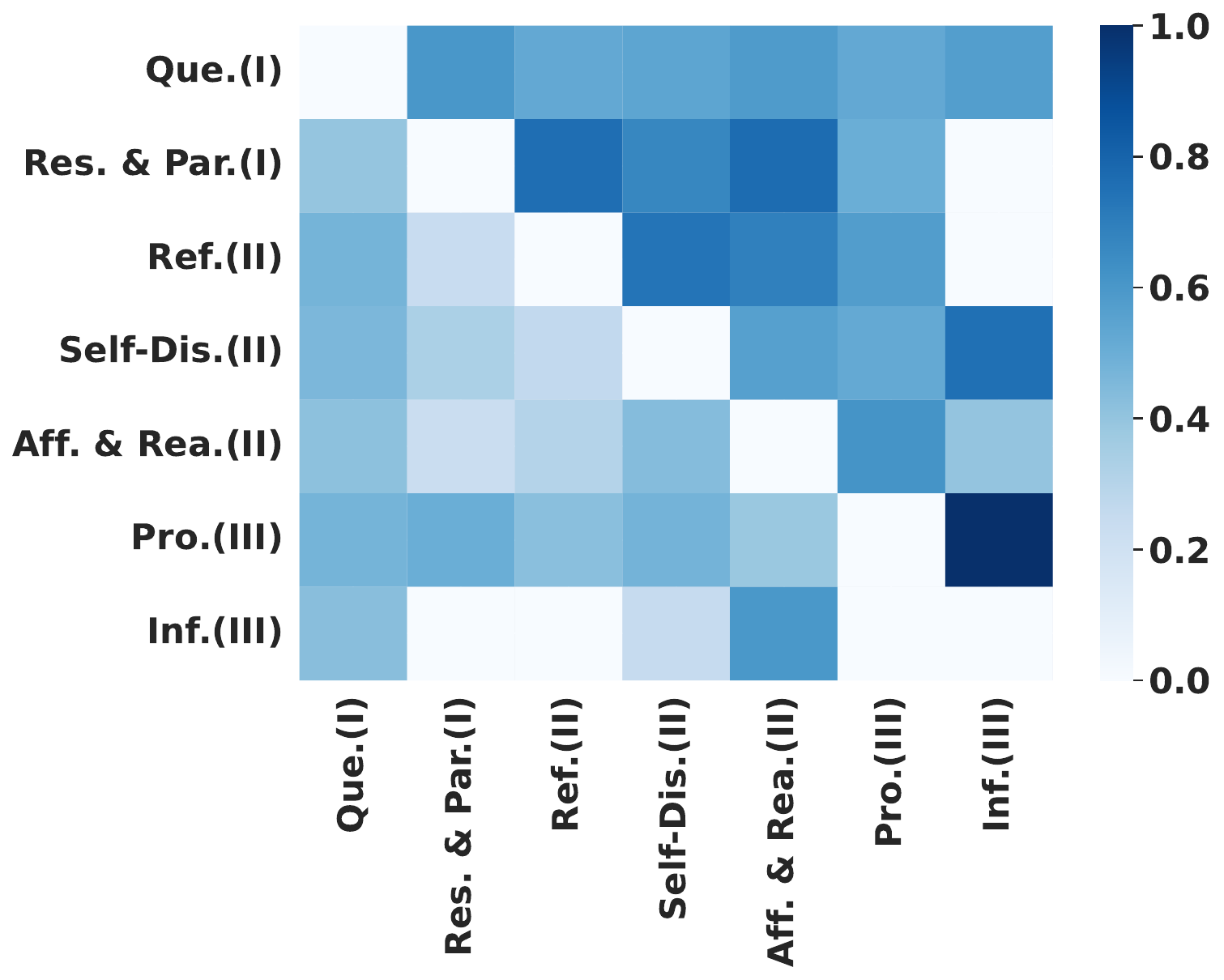}
    \hspace{0.1in}
    \includegraphics[width=0.48\linewidth]{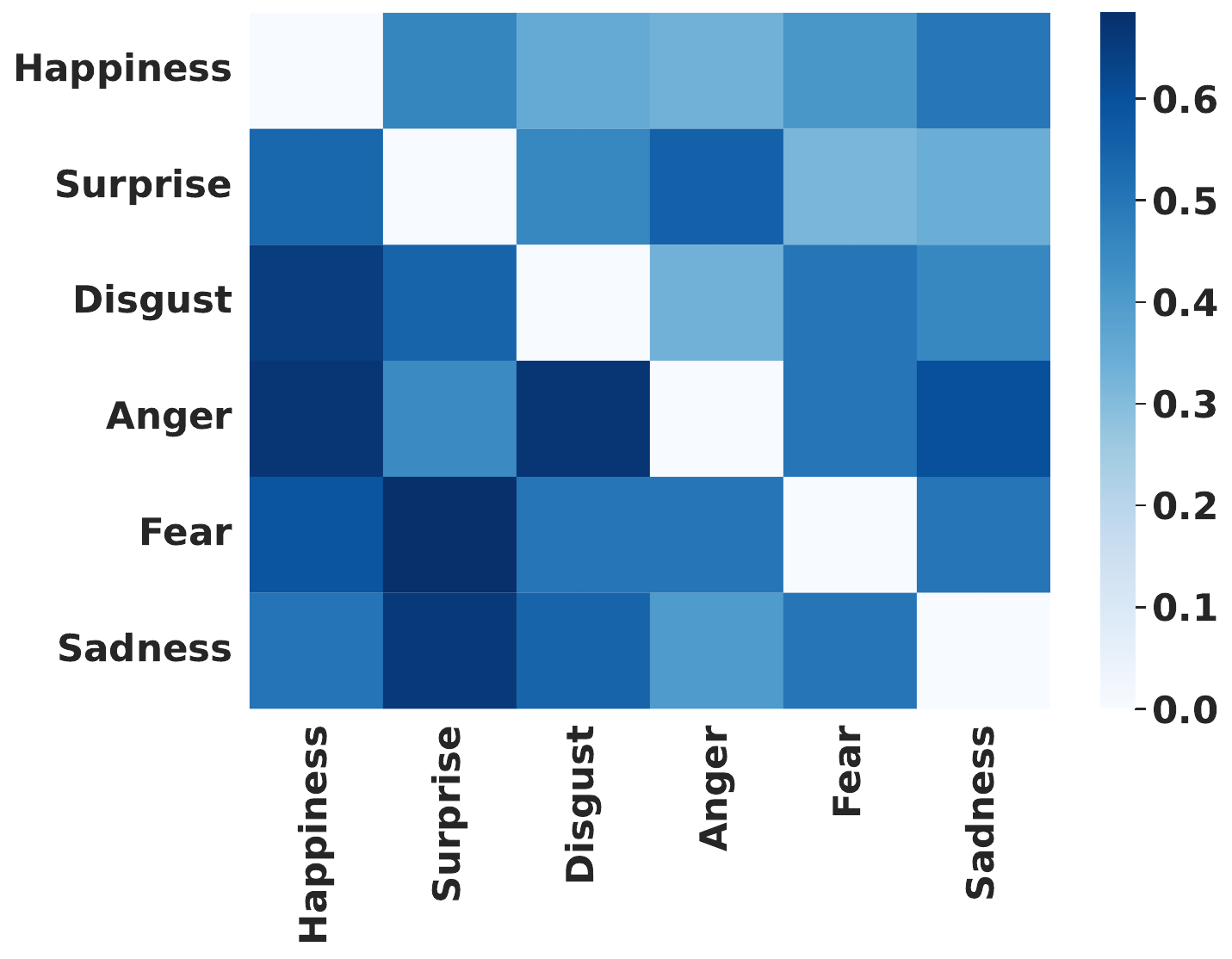}
    %\vspace{-3.5mm}
    \caption{Distribution of conceptual transitions within the utterance. Left: the strategy transition on ESConv. Right: the emotion transition on DailyDialogue.}
    \label{fig:concept_transition}
\end{figure*}

\paragraph{Conceptual transition within utterance.} Compared to previous approaches, {\ModelName} allows for the transitions between different concepts (\textit{e.g.}, recognition of emotions and strategies) within a single utterance, which is aligned with the human-like thinking behavior, and closer to the original annotated response of ESConv. To further illustrate this behavior, we exhibit the distribution of the inner-utterance conceptual transitions in Figure \ref{fig:concept_transition}. For a specific grid $(i, j)$,  $i$ denotes the row index and $j$ denotes the column index, and its color level indicates the relative occurrence of transition $c_i \rightarrow c_j$.

%Figure: \textcolor{red}{difference between inner-utterance transition and outer-utterance transition of strategies}

Figure \ref{fig:concept_transition} (left) exhibits the strategy transitions generated by {\ModelName} on ESConv. According to \cite{liu2021ESConv}, the strategies can be related to three supporting stages (Exploration, Comforting and Action), therefore we mark each strategy with the corresponding stage I, II and III and order the strategies accordingly. As a result, a more reasonable transition may happen from a left strategy to a right one, \textit{i.e.}, the upper-triangle part of the matrix. Obviously, such a pattern can be observed from Figure \ref{fig:concept_transition} (left), indicating that the strategy determination of {\ModelName} is aligned with the `Exploration, Comforting and Action' paradigm.
%For each grid, the color represents the fraction of the transition between the row-strategy to the column-strategy. 

Figure \ref{fig:concept_transition} (right) shows the emotion transitions generated by {\ModelName} on DailyDialog. From the distribution, we can observe that more frequent emotion transitions of {\ModelName} happen on Disgust\&Anger $\rightarrow$ Happiness, Anger $\rightarrow$ Disgust, and Fear\&Sadness $\rightarrow$ Surprise. On the contrary, the over-extreme emotional transition (Happiness $\rightarrow$ Sadness) has a low frequency. {\ModelName} mimics the human emotion curve when expressing opinions, providing a more fine-grained emotional interaction.
% Anger $\rightarrow$ Fear, Fear $\rightarrow$ Guilt, and Disgust $\rightarrow$ Depression

\paragraph{{\ModelName} without pre-defined concepts.} Previous results (especially OOD) suggest that the pre-defined concepts may hinder {\ModelName} from exploring broader scopes. In Table \ref{tab:ablation_results}, we also try to allow the LLM to generate concepts on-the-fly (\textit{free concept}). Surprisingly, although \textit{free concept} endures some performance degradation, it still performs well and surpasses the other baselines. This observation suggests the potential of {\ModelName} as a more generalized methodology, \textit{e.g.}, remove the concept list in the prompt, but pretrain the LLM with some conceptual knowledge.

\begin{table}[t!]
\caption{\textbf{The finetuning result of {\ModelName} on LLaMA3-70B-Instruct, from responses of GPT4o.}}
\label{tab:finetuning}
\renewcommand{\arraystretch}{1.11} 
\centering
\small
% \resizebox{1.8\columnwidth}{!}{
\begin{tabular}{l | cccc } 
    \toprule
    \multirow{2}{*}{Method $\downarrow$} &  
    \multicolumn{4}{c}{ESConv} \\ 
    \cmidrule{2-5} & B-2 & R-L & D-2 & CDr \\ 
    \midrule 
    %LLaMA3 (8B) \\  
    %\midrule
    %\;+ \ModelName       & \bf4.75 & 13.31 & 18.54 & 4.87 \\ 
    %\;+ \ModelName  w/ finetuning      & 4.61 & \bf13.60 & \bf45.34 & \bf7.49 \\ 
    %\midrule
    %LLaMA3 (70B) \\
    \ModelName        & 4.30 & 12.71 & 18.16 & 2.07  \\ 
    \ModelName  w/ finetuning      & \bf 4.75 & \bf 13.49 & \bf 42.54 & \bf 8.68 \\ 
    \bottomrule
\end{tabular}
\end{table}

\paragraph{{\ModelName} with finetuning.} Although {\ModelName} is primarily proposed as a prompt-based paradigm, it is also possible to collect the {\ModelName} result from a stronger LLM, then finetune it on a smaller LLM. This methodology is also known as distillation from the teacher LLM. In Table \ref{tab:finetuning}, we compare the 70B result of {\ModelName} with finetuning to the original result, indicating that {\ModelName} is friendly to finetuning, and its performance can be further enhanced by distillation from a teacher, \textit{i.e.}, GPT4o.

%CoCT yields inherent performance enhancements relative to baseline, which serves to validate the design rationality of CoCT for emotional support tasks. Building upon this foundation, task-specific finetuning acts as a performance amplifier, further augmenting the advantages conferred by CoCT. As shown in Table \ref{tab:finetuning} , metrics indicative of response diversity (e.g., Dist-2) and semantic quality (e.g., CIDEr) exhibit substantial improvements following finetuning. For instance, in the case of LLaMA3-8B, the Dist-2 metric surges from 18.54 to 45.34 after finetuning. Such gains facilitate the generation of responses that are not only more diverse but also semantically aligned with the nuances of emotional support scenarios. 

\paragraph{Choice of special tokens.} The prompt of {\ModelName} requires an arbitrary choice of the special token to tag the concept term. In Section \label{sec:method}, we take `<>' as the example, while here we compare its results to other special tokens (e.g., \#, @, \^, [], \&). Table \ref{tab:special_tokens} exhibits this comparison result on 8B and 79B, showing that the current choice is optimal on 70B, while different special tokens may perform similarly on 8B. In practice, one may conduct similar tests to determine the special token before the formal experiment.

% Table \ref{tab:special_tokens} assesses the impact of distinct special tokens on the performance of LLaMA3-8B and LLaMA3-70B across ESConv metrics. Here, special tokens refer to custom symbols (e.g., <> , \# , @ ,\^, [], \&) introduced to guide model behavior in conversations. For LLaMA3-70B, '<>' exhibits notable strengths. While its B-2 (3.96) and R-L (12.14) are competitive, it shows a more balanced performance across metrics.

\begin{table}[t!]
    \caption{\textbf{Performance comparison of different special tokens on ESConv.}} 
    \label{tab:special_tokens}
    \renewcommand{\arraystretch}{1.11} 
    \centering
    \small
    \begin{tabular}{l | cccc }
        \toprule
        \multirow{2}{*}{Method $\downarrow$} &  
        \multicolumn{4}{c}{ESConv} \\  
        \cmidrule{2-5} 
        & B-2 & R-L & D-2 & CDr \\  
        \midrule   
        LLaMA3 (8B) \\ 
        \;w/ special token: <>       & 4.75 & 13.31 & 18.54 & 4.87 \\ 
        \;w/ special token: \^        & 4.38 & 13.16 & 18.34 & 3.79 \\ 
        \;w/ special token: \#       & \bf 4.86 & \underline{14.12} & 18.65 & 4.99 \\  
        \;w/ special token: @        & 4.54 & 13.54 & 18.43 & 4.07 \\ 
        \;w/ special token: []       & \underline{4.82} & \bf 14.13 & \underline{18.78} & \bf 5.09 \\ 
        \;w/ special token: \&       & 4.72 & 13.71 & \bf 19.04 & 4.62 \\  
        \midrule
        LLaMA3 (70B) \\  
        \;w/ special token: <>       & \bf 4.30 & \bf 12.71 & \bf 18.16 & \bf2.07 \\ 
        \;w/ special token: \^        & 3.96 & 12.14 & 16.60 & 1.32 \\ 
        \;w/ special token: \#       & \underline{4.17} & \underline{12.68} & \underline{17.04} & \underline{1.88} \\  
        \;w/ special token: @        & 4.11 & 12.54 & 16.68 & 1.74 \\ 
        \;w/ special token: []       & 4.01 & 12.45 & 16.18 & 1.48 \\ 
        \;w/ special token: \&       & 3.92 & 12.06 & 15.94 & 1.25 \\  
        \bottomrule
    \end{tabular}  
\end{table}

%\paragraph{Limitation and bad case analysis.} 

%\subsection{Sensitivity Analysis}

\section{Related Work}
\label{sec:related_work}

%\subsection{Prompt-based Studies}

Large language models can generalize to different tasks given an informative and instructional prompt. Chain-of-thought (CoT) \cite{wei2022chain} is a famous study that prompts the LLM by the paraphrase `Let's think step by step' and obtains substantial performance improvement on metathetical and reasoning tasks. Further studies are proposed to enhance the performance by using different prompt paradigms. For example, Chain-of-Hindsight (CoH) \cite{liu2024CoH} allows the LLM to reflect from human feedback and have a better alignment with human preference. Self-Refine \cite{Madaan2023SelfRefine} let the LLM generate feedback from its initial response, then refine the response again based on this feedback. SoT \cite{ning2024skeletonofthought} first guides LLMs to generate the skeleton of the answer, and then conducts parallel API calls or batched decoding to complete the contents.

In contrast, here we mainly focus on the open-ended conversation scenario and argue that LLM can also benefit from a chain of concepts, instead of reasoning steps. Such concepts may include emotions, strategies and topics, and can transit within the utterance. LLM is prompted to first generate the sketchpad of concepts, followed by the formal response. %Our experiment verifies that our method is more effective than CoT on non-reasoning tasks, such as Open-ended conversations.

\section{Conclusion}
\label{sec:conclusion}

In this work, we propose a novel prompt paradigm called {\ModelName} which can be applied to open-domain conversations. Motivated by humans' conceptual ideas such as emotions, strategies and topics, we let the LLM first generate the tag of the concept, then produce the detailed content of the corresponding concept. This chain of conceptual thoughts facilitates the LLM to have a hierarchical and strategic way of thinking, highlighting its decision transitions within a single utterance. Experiments on daily and emotional support conversations show that {\ModelName} produces better responses than other prompt-based baselines, given either in-domain and out-of-domain concepts and queries. Instead of providing a pre-defined concept list, {\ModelName} can also generate the answer with concepts generated on-the-fly, potentially implying a more generalized generative paradigm.

%\clearpage
%\newpage

\section{Limitation}
\label{sec:limitation}

{\ModelName} is orthogonal to few-shot demonstrations, finetuning, self-reflection, and usage of structured or external knowledge. Combinations of these methods might further benefit.

Currently, {\ModelName} also needs a manually provided list of concept candidates, which might constrain its application. Future work may consider invoking the LLM's hierarchical thinking mechanism, which automatically first formulates the next key concept, then deduces the detailed response.

Although typical cases and quantitative metrics primarily indicate the effectiveness of {\ModelName}, due to the capability limitation of the foundation model, {\ModelName} still sometimes fails to understand the concept list, or conduct the semantically wrong conceptual transition. We also include a bad case analysis in the Appendix.

\section{Ethical Considerations}

{\ModelName} can exhibit inconvenient, aggressive behaviors given specific concepts. It is suggested that the concept list needs to be carefully checked before the formal deployment. If concepts are generated on-the-fly, the finetuning may be needed to constrain the behavior within safety.

\iffalse
\section*{Acknowledgments}

This document has been adapted
by Steven Bethard, Ryan Cotterell and Rui Yan
from the instructions for earlier ACL and NAACL proceedings, including those for
ACL 2019 by Douwe Kiela and Ivan Vuli\'{c},
NAACL 2019 by Stephanie Lukin and Alla Roskovskaya,
ACL 2018 by Shay Cohen, Kevin Gimpel, and Wei Lu,
NAACL 2018 by Margaret Mitchell and Stephanie Lukin,
Bib\TeX{} suggestions for (NA)ACL 2017/2018 from Jason Eisner,
ACL 2017 by Dan Gildea and Min-Yen Kan,
NAACL 2017 by Margaret Mitchell,
ACL 2012 by Maggie Li and Michael White,
ACL 2010 by Jing-Shin Chang and Philipp Koehn,
ACL 2008 by Johanna D. Moore, Simone Teufel, James Allan, and Sadaoki Furui,
ACL 2005 by Hwee Tou Ng and Kemal Oflazer,
ACL 2002 by Eugene Charniak and Dekang Lin,
and earlier ACL and EACL formats written by several people, including
John Chen, Henry S. Thompson and Donald Walker.
Additional elements were taken from the formatting instructions of the \emph{International Joint Conference on Artificial Intelligence} and the \emph{Conference on Computer Vision and Pattern Recognition}.
\fi

\bibliography{custom}

%\newpage

\appendix

\section{More Dataset Details}
\label{appendix:dataset}

\paragraph{Detailed introductions.} We introduce our experimental datasets in more detail;
\begin{itemize}
    \item The \textit{DailyDialogue} dataset \cite{li-etal-2017-dailydialog} is  a widely used benchmark for Open-ended conversation. It contains 13,118 multi-turn dialogues across diverse daily topics (e.g., family, work, hobbies) with manually annotated emotions and dialogue strategies (named `act' in the original dataset). It is annotated with 7 emotions and 4 strategies.

    \item The \textit{MultiWoZ} dataset \cite{budzianowski-etal-2018-multiwoz} is a task-oriented dialogue corpus focused on service-oriented interactions (e.g., hotel booking, restaurant reservations, transportation inquiries), with no concept annotations.

    \item The \textit{ESConv} dataset \cite{liu2021ESconv} is a specialized corpus for emotional support research. It contains 1,000+ multi-turn dialogues where users articulate personal struggles (e.g., workplace stress, interpersonal conflicts, self-esteem issues), and support providers respond with empathy, validation, and actionable guidance. It is annotated with 11 emotions and 8 strategies.

    \item The \textit{EmpatheticDialogues} dataset \cite{rashkin-etal-2019-towards} contains conversations centered on recognizing and responding to emotions (e.g., joy, sadness, frustration). It has annotations of user emotions but no strategies.
\end{itemize}

\paragraph{Dataset statistics.} Table \ref{tab:statistics} summarizes the statistics of ESConv, DailyDialogue, and EmpatheticDialogues. Table \ref{tab:ESConv_strategies} provides definitions of support strategies in ESConv and DailyDialogue. 

\begin{table*}[ht!] 
  \caption{Statistics of ESConv, DailyDialogue, EmpatheticDialogues. %For ESConv, we removed utterances from supporters at the beginning of dialogues because these utterances are usually uninformative greetings.
  }
  \label{tab:statistics}%
  
  \centering
  \small
  %\resizebox{0.98\textwidth}{!}{
    \begin{tabular}{llccc}
    \toprule
    %\multicolumn{3}{c}{\textbf{Statistics of ESConv}} \\ 
    \multicolumn{2}{c}{Category} & ESconv & DailyDialogue & EmpatheticDialogues   \\
    \midrule
    \multicolumn{2}{l}{\# Sessions} & 1.3K & 13.1k& 2.5K \\
    %\multicolumn{2}{l}{Average Session Length} & 543.6  \\
    \multicolumn{2}{l}{\# Utterances} & 38K & 103.0k& 11.0K\\
    \multicolumn{2}{l}{Average \# Utterances} & 28.9  & 7.9& 4.3\\
    \multicolumn{2}{l}{Average Utterance Length} & 18.8  & 13.6& 16.7\\
    \midrule
    \multirow{5}[0]{*}{Seeker/Speaker1} & \# Utterances & 20K&53.8k& 5.7K \\
       & Avg \# Utterances & 15.4 & 4.1& 2.2 \\
       & Avg Uttr Len & 16.8& 13.2& 20.8  \\
        & \# Strategies & -& 4& -\\
       & \# Emotions & 11& 7& 32 \\
    \midrule
    \multirow{5}[0]{*}{Supporter/Speaker2} & \# Utterances & 18K & 49.2k& 5.2K\\
       & Avg \# Utterances & 13.6& 3.9& 2.1  \\
       & Avg Uttr Len & 21.0& 14.1& 12.3  \\
       & \# Strategies & 8& 4& -\\
       & \# Emotions & -& 7& 32 \\
    \bottomrule
    \end{tabular}%
  %}

\end{table*}

\begin{table*}[h!]
\caption{Strategy definitions and abbreviations of ESConv and DailyDialogue.}
\centering
\small
\setlength{\tabcolsep}{8pt} % 调整列间距
\resizebox{0.98\textwidth}{!}{
\begin{tabular}{>{\centering\arraybackslash}m{0.1\textwidth}m{0.25\textwidth}m{0.1\textwidth}m{0.65\textwidth}}
\Xhline{1.2pt}
\multicolumn{1}{l}{\textbf{Dataset}} &\multicolumn{1}{l}{\textbf{Strategies}} & \multicolumn{1}{c}{\textbf{Abbreviation}} &\multicolumn{1}{c}{\textbf{Definitions}}  \\ 
\toprule
\multirow{14}{*}{\textbf{ESconv}}&Question & Que.& Inquiring about problem-related information to help the seeker clarify their issues, using open-ended questions for best results and closed questions for specific details. \\ 
&Restatement or Paraphrasing &Res.\& Par.& A simple, more concise rephrasing of the help-seeker’s statements that could help them see their situation more clearly. \\ 
&Reflection of Feelings &Ref.& Articulate and describe the help-seeker’s feelings. \\ 
&Self-disclosure &Self-Dis.& Divulge similar experiences that you have had or emotions that you share with the help-seeker to express your empathy. \\ 
&Affirmation and Reassurance &Aff. \& Rea.& Affirm the help seeker’s strengths, motivation, and capabilities and provide reassurance and encouragement. \\
&Providing Suggestions &Pro.& Provide suggestions about how to change, but be careful to not overstep and tell them what to do. \\ 
&Information & Inf.& Provide useful information to the help-seeker, for example with data, facts, opinions, resources, or by answering questions. \\ 
&Others &others& Exchange pleasantries and use other support strategies that do not fall into the above categories. \\ 
\toprule
\multirow{7}{*}{\textbf{DailyDialogue}}
&Inform &-& Provide factual or contextual information that the speaker believes the listener may not know or is unaware of. \\ 
&Question &-.& Seek specific information from the listener, assuming they possess the knowledge being requested. \\ 
&Directives &-& Express the speaker’s intention for the listener to take an action, including requests, instructions, or suggestions. \\ 
&Commissive &-& Indicate the speaker’s commitment to perform certain actions, such as accepting or rejecting requests or offers. \\
\Xhline{1.2pt}
\end{tabular}
}
\label{tab:ESConv_strategies}
\end{table*}

\section{More Evaluation Details}
%\subsection{Evaluation Details}
\label{appendix:evaulation}

\paragraph{Automatic Metrics.}

\textbf{Bleu-2}(B-2)\citep{papineni2002bleu} first compute the geometric average of the modified $n$-gram precisions, $p_n$, using $n$-grams up to length $N$ and positive weights $w_n$ summing to one. Next, let $c$ be the length of the prediction and $r$ be the reference length. The BP and Bleu-2 are computed as follows.

\begin{align}
    \mathrm{BP}=\left\{\begin{array}{ll}
1 & \text { if } c>r \\
e^{(1-r / c)} & \text { if } c \leq r
\end{array} .\right. \\
    \mathrm{Bleu}=\mathrm{BP} \cdot \exp \left(\sum_{n=1}^N w_n \log p_n\right) .
\end{align}

\textbf{Rouge-L (R-L)}\citep{lin2004rouge} propose using LCS-based F-measure to estimate the similarity between two summaries $X$ of length $m$ and $Y$ of length $n$, assuming $X$ is a reference summary sentence and $Y$ is a candidate summary sentence, as follows:

\begin{equation}
\begin{aligned}
& R_{l c s}=\frac{L C S(X, Y)}{m} \\
& P_{l c s}=\frac{L C S(X, Y)}{n} \\
& F_{l c s}=\frac{\left(1+\beta^2\right) R_{l c s} P_{l c s}}{R_{l c s}+\beta^2 P_{l c s}}
\end{aligned}
\label{rouge_l}
\end{equation}

Where $\operatorname{LCS}(X, Y)$ is the length of a longest common subsequence of $X$ and $Y$, and $\beta=P_{l c s} / R_{\text {lcs }}$ when $\partial F_{l c s} / \partial R_{l c s}=\partial F_{l c s} / \partial P_{l c s}$. In DUC, $\beta$ is set to a very big number $(\rightarrow \infty)$. Therefore, the LCS-based F-measure, i.e. Equation \ref{rouge_l}, is Rouge-L. 

\textbf{CIDEr} \citep{vedantam2015cider} score for $n$-grams of length $n$ is computed using the average cosine similarity between the candidate sentence and the reference sentences, which accounts for both precision and recall:
\begin{equation}\label{eq:2}
CIDEr_n(c_i, S_i) = \frac{1}{m}\sum_j \frac{\textbf{g}^\textbf{n}(c_{i})\cdot \textbf{g}^\textbf{n}(s_{ij})}{\|\textbf{g}^\textbf{n}(c_{i})\|\|\textbf{g}^\textbf{n}(s_{ij})\|},
\end{equation}
where $\textbf{g}^\textbf{n}(c_{i})$ is a vector formed by $g_k(c_{i})$ corresponding to all $n$-grams\ of length $n$ and $\|\textbf{g}^\textbf{n}(c_{i})\|$ is the magnitude of the vector $\textbf{g}^\textbf{n}(c_{i})$. Similarly for $\textbf{g}^\textbf{n}(s_{ij})$.

Higher order (longer) $n$-grams are used to capture grammatical properties as well as richer semantics. \citep{vedantam2015cider} combine the scores from $n$-grams of varying lengths as follows:

\begin{equation} \label{eq:3}
CIDEr(c_i, S_i) = \sum_{n=1}^N w_n CIDEr_n(c_i, S_i),
\end{equation}
Empirically, Vedantam et al.\citep{vedantam2015cider} found that uniform weights $w_n=1/N$ work the best. So, We also use $N$ = 4.

\paragraph{LLM-as-a-Judge.}

Table \ref{tab:prompt_comparison_score} provides the detailed prompt we feed to GPT4o for comparison evaluation.

\begin{table*}[t!]
\caption{The Comparison Evaluation Prompt Content by GPT-4o from \cite{madani2024steeringconversationallargelanguage}}
\label{tab:prompt_comparison_score}
\centering
\small
\resizebox{0.99\textwidth}{!}{
\begin{tabular}{p{17cm}}
    \toprule[1.5pt]
    \textbf{Prompt}\\ 
    \toprule
    
    Please act as an impartial judge and evaluate the quality of the responses provided by two AI assistants to the user questions. You should choose the assistant that follows the provided strategy more carefully and precisely to answer the user's last utterance. You should use your emotional support expertise and knowledge to judge the quality of the response considering how well the answer follows the provided strategy. Your evaluation most importantly should consider strategy adherence and then the overall quality, naturalness, consistency and coherence of the final utterance.\\
    \\
	Begin your evaluation by comparing the responses of the two assistants and provide a short explanation. Avoid any position biases and ensure that the order in which the responses were presented does not influence your decision. Do not allow the length of the responses to influence your evaluation. Do not favor certain names of the assistants. Be as objective as possible. After providing your explanation, output your final verdict by strictly following this format: "JUDGE: [[A]]" if assistant A is better, "JUDGE: [[B]]," if assistant B is better, and "JUDGE: [[C]]" for a tie.\\
    \\
	Conversation history:\\
    \\
    \{conversation\_history\}\\
    \\
    <|The Start of Assistant A's Response|>  \\
    \\
    \{assistant\_a\_resp\}  \\
    \\
    <|The End of Assistant A's Response|> \\
    \\
    <|The Start of Assistant B's Response|>  \\
    \\
    \{assistant\_b\_resp\}  \\
    \\
    <|The End of Assistant B's Response|> \\
    \bottomrule[1.5pt]
\end{tabular}
}
\end{table*}

\paragraph{Human evaluation criteria.}

We enlisted four interns as human evaluators to rate the models across multiple dimensions, specifically Fluency, Emotion, and Satisfaction. The Satisfaction dimension encompasses four sub-aspects: Acceptance, Effectiveness, Sensitivity, and Alignment. Throughout this evaluation, we strictly adhered to international regulations and ethical norms, ensuring that all procedures comply fully with the guidelines regarding participant participation and data integrity. Each evaluator was required to assess every sample independently and strictly in accordance with the pre - pre-established criteria. By upholding these principles, the evaluation process maintained objectivity, standardization, and consistency, thereby improving the overall quality and credibility of the evaluation results.

The detailed manual scoring criteria are as follows:
\begin{itemize}
\item Fluency:

1: The sentence is highly incoherent, making it extremely difficult to understand and failing to convey a meaningful idea.

2: The sentence has significant incoherence issues, with only parts of it making sense and struggling to form a complete thought.

3: The sentence contains some incoherence and occasional errors, but can still convey the general meaning to a certain extent.

4: The sentence is mostly fluent with only minor errors or slight awkwardness in expression, and effectively communicates the intended meaning.

5: Perfect. The sentence is completely fluent, free of any errors in grammar, punctuation, or expression, and clearly conveys the idea.

\item Emotion:

1: The emotional expression is extremely inappropriate and chaotic, not in line with the content, and may convey wrong emotions.

2: The emotional expression has obvious flaws, either too weak or exaggerated, and is disjointed from the content.

3: The emotional expression is average. It can convey basic emotions but lacks depth and has minor issues.

4: The emotional expression is good. It can effectively convey the intended emotion with an appropriate intensity and is well integrated with the content.

5: The emotional expression is excellent. It is rich, nuanced, and perfectly matches the content, capable of evoking a strong and appropriate emotional response.

\item Acceptance:

1: The response inescapably triggers emotional resistance.

2: The response is highly likely to trigger emotional resistance.

3: The response has a possibility of emotional resistance occurring.

4: The response rarely provokes emotional resistance.

5: The response has no occurrence of emotional resistance.

\item Effectiveness:

1:  The response actually worsens the seeker's emotional distress.

2: The response carries the risk of increasing stress levels, and this outcome varies depending on the individual user.

3: The response fails to alter the seeker's current emotional intensity and keeps it at the same level.

4: The response shows promise in calming the emotional intensity; however, it is overly complicated or ambiguous for the user to fully comprehend and utilize effectively.

5: The response appears to be highly effective in soothing the seeker's emotions and offers valuable and practical emotional support. 

\item Sensitivity:

1: The response renders inaccurate evaluations regarding the seeker's state.

2: The response is characterized by rash judgments, as it lacks adequate assessment and in-depth exploration of the seeker's state.

3: The response is formulated with a one-sided judgment and a limited exploration of the seeker's state.

4: The response demonstrates an understanding that only covers a part of the seeker's state.

5: The response precisely grasps the seeker's state and is appropriately tailored according to the seeker's actual situation.

\item Alignment:

1: The response is in total contradiction to the predicted strategy.

2: The response has a minor deviation from the predicted strategy.

3: There is some ambiguity between the response and the predicted strategy.

4: The response largely matches the predicted strategy, yet it contains some ambiguous elements.

5: The response effectively makes itself consistent with the predicted strategy.

\item Satisfaction:

1: The response is extremely disappointing. It doesn't answer the question at all and is of no help.

2: The response is poor. It only gives a partial answer and leaves many doubts unresolved.

3: The response is average. It meets the basic requirements but isn't particularly outstanding.

4: The response is good. It answers the question clearly and provides some useful details.

5: The response is excellent. It not only answers the question perfectly but also offers valuable additional insights.

\end{itemize}

%four annotators are employed to provide quality scoring. 
At the end, the cross-validation is conducted with the averaged results collected. 

\section{Detailed Introductions of Baselines}
\label{appendix:baseline}

We introduce the baselines with more details: 
%\textbf{\textcolor{red}{add intro of more baselines @guqingqing(done)}}

\begin{itemize}
    \item Direct: directly inference the LLM.
    \item Direct-Refine: a straightforward refinement method in which the model revises its initial response to incorporate emotional support considerations.
    \item Self-Refine \cite{Madaan2023SelfRefine}: generates feedback about the emotional support based upon its initial response, then refines its response based on this feedback.
    \item ECoT \cite{li2024enhancingemotionalgenerationcapability}: an analogy of Chain-To-Thought prompt \cite{wei2022chain} on the emotional conversation, which first generates the seeker's \textit{emotion}, which then guides the generation of strategy and response.
    \item SoT \cite{ning2024skeletonofthought}: first guides LLMs to generate the skeleton of the answer, and then in parallel decodes to complete the contents.% of each skeleton point in parallel. 
    \item ToT  \cite{yao2023tree}: decomposes complex problems into intermediate thought steps (like a tree structure), where LLMs evaluate the validity of each step and iteratively explore multiple reasoning paths to select the optimal solution, enhancing the ability to handle multi-step logical tasks. 
    \item Plan-and-Solve \cite{wang2023planandsolvepromptingimprovingzeroshot}: first prompts LLMs to generate a detailed plan outlining sub-goals and reasoning strategies, then executes the plan step-by-step to complete the solution, integrating planning and execution to improve the coherence and completeness of responses, especially in mathematical reasoning and multi-turn decision scenarios.
    \item RAG \cite{fan2024RAGmeetingLLMs}: the retrieval-augmented generation, which chooses the appropriate strategy by a semantic retriever, and includes it in the prompt to improve the response quality. In this work, we employ E5-Large \cite{wang2024E5} as the retriever.
    %\item SFT: finetuning on the training sets, including the strategy and emotion labels.
    \item CSIM \cite{zhou-etal-2024-think}: a self-chat method which produces the response based on its implicit simulation of future dialogue. 
\end{itemize}

\section{More Experimental Results}
\label{appendix:experiment}

\subsection{More Typical Cases}

Table \ref{tab:case_dailydialogue} provides typical cases in DailyDialogue. From the contents generated, we can observe that {\ModelName} can recognize the concept list, generate a chain of concepts, and then produce the detailed response based on the concept skeleton. The concept flow induces the LLM's reasoning path. Table \ref{tab:multiwoz_case} shows cases of MultiWoz which show that {\ModelName} adapts well to MultiWoz's topics.

\begin{table*}[h!]
    \caption{Cases of {\ModelName} in DailyDialogue, in contrast to Direct Inference.} 
    \label{tab:case_dailydialogue}
    \centering
    \small
    %\resizebox{\textwidth}{!}{%
    \begin{tabular}{c|c|l}
        \toprule
        %\multicolumn{2}{c}{\textbf{Case 1}}  \\ 
        %\toprule
        \multicolumn{1}{c|}{\multirow{3}[2]{*}{\rotatebox[origin=c]{90}{\centering Case 1}}} 
        & user & \makecell[l]{\textit{I'm sorry I'm so late . I had a really bad day.}} \\
        \cmidrule{2-3}
        & Direct & \makecell[l] {It's ten after six. We're late. But dinner is at six thirty.} \\
        \cmidrule{2-3}
        & CoCT & \makecell[l]{\textbf{<Commissive>} No need to apologize; I'm just glad you're here now. \\\textbf{<Question>} Do you want to talk about your day? Sometimes sharing can\\ really help. } \\
        %\bottomrule
        %\multicolumn{2}{c}{\textbf{Case 2}}  \\ 
        \toprule
        \multicolumn{1}{c|}{\multirow{3}[2]{*}{\rotatebox[origin=c]{90}{\centering Case 2}}} 
        & user & \makecell[l]{\textit{But Sunday is the same with Saturday. If you don’t mind, that's OK.}}    \\ 
        \cmidrule{2-3}
        & Direct & \makecell[l]{Hum.Maybe it will be better. The most important thing is that I can have\\ a day to relax after a week's work .}   \\
        \cmidrule{2-3}
        & CoCT & \makecell[l]{\textbf{<Surprise>} Oh, really? I never noticed that.\\ \textbf{<Question>} Are you sure Sunday is just as busy as Saturday? \\Maybe we could go early in the morning when it's less crowded?}   \\
        %\toprule
        %\multicolumn{2}{c}{\textbf{Case 3}}  \\ 
        \toprule
        \multicolumn{1}{c|}{\multirow{3}[2]{*}{\rotatebox[origin=c]{90}{\centering Case 3}}} 
        & user & \makecell[l]{\textit{Believe it or not, tea is the most popular beverage in the world after water.}} \\
        \cmidrule{2-3}
        & Direct & \makecell[l]{Well, people from Asia to Europe all enjoy tea.}   \\
        \cmidrule{2-3}
        & CoCT & \makecell[l]{\textbf{<Inform>} Yes, that's true! Tea is indeed the most widely consumed\\ beverage in the world after water. Its popularity spans across many cultures\\ and countries, with different varieties like black, green, white, and oolong tea.}   \\
    \bottomrule
    \end{tabular}
    %}
\end{table*}

\begin{table*}[h!]
    \caption{Case of {\ModelName} on MultiWOZ, in contrast to the original response.} 
    \label{tab:multiwoz_case}
    \centering
    \small
    %\resizebox{\textwidth}{!}{%
    \begin{tabular}{c|l}
        \Xhline{2\arrayrulewidth}
        %\multicolumn{2}{c}{\textbf{w/ emotion+strategy}}  \\ 
        %\hline
        user & \makecell[l]{\textit{I will be needing a place to stay on the west side of town.}}  \\
        \hline
        assistant & \makecell[l]{Hobsons house is in the west part of town, would you like to stay there?} \\ 
        \hline
        CoCT & \makecell[l]{\textbf{<Affirmation and Reassurance>} You've got a good head start on\\ planning your trip to Cambridge. \textbf{<travel>} For accommodations on the\\ west side. I can suggest checking out hotels near the River Cam or universities\\ like Churchill College, what's your budget like for the stay?}   \\

        %\hline
        %seeker & \makecell[l]{\textit{I put too much sauce on it, so it ended up a little salty.}}  \\
        %\hline
        %supporter & \makecell[l]{<Reflection of feelings> I can understand how that might have been disappointing, \\ especially when you were looking forward to enjoying your homemade pizza. \\ < Self-disclosure> I once did the same thing with a pasta dish - it was so salty! \\ <Providing Suggestions and Information> Next time, you might try adding less sauce and then taste-testing as you go. \\ Cooking is full of little adjustments and learning experiences. \\ <Questioning> What other recipes are you thinking about trying?}  \\
        \Xhline{2\arrayrulewidth}
    \end{tabular}
    %}
\end{table*}
\begin{table*}[h!]
    \caption{Bad case example of {\ModelName} in ESconv.} 
    \label{tab:case_bad}
    \centering
    \small
    %\resizebox{\textwidth}{!}{%
    \begin{tabular}{c|l}
        %\toprule
        %\multicolumn{2}{c}{\textbf{Bad case 1}}  \\ 
        \toprule
        user & \makecell[l]{\textit{I just need ways to find validation that are not based on my job. But validation doesn't pay the bills.}} \\ % Yeah. I mean I know a lot of people struggle the same way. 
        \midrule
        {\ModelName} & \makecell[l]{\textbf{<Reflection of feelings>}  It sounds like you're feeling a bit lost and uncertain about your sense of \\ purpose and identity outside of your job. \\ \textbf{<Questioning>} Can you tell me more about what you mean by "validation", what does that look like \\for you, and what are some things that have given you a sense of fulfillment and self-worth in the past? \\ \textcolor{red}{\textbf{<Hobbies and crafts>} Perhaps we can explore some hobbies or activities that bring you joy and help} \\ \textcolor{red}{ you feel more connected to yourself and others.}} \\
    \bottomrule
    \end{tabular}
    %}
\end{table*}

\subsection{Bad Case Analysis}

Table \ref{tab:case_bad} provides a typical bad case. The red part indicates that {\ModelName} attempts to transfer to the topic of `Hobbies and crafts', however, which is almost the same as the original topic. Improvement of foundation models may alleviate such issues.

\end{document}